\documentclass{article}

\usepackage{arxiv}

\usepackage[utf8]{inputenc} 
\usepackage[T1]{fontenc}    
\usepackage{hyperref}       
\usepackage{url}            
\usepackage{booktabs}       
\usepackage{amsfonts}       
\usepackage{nicefrac}       
\usepackage{microtype}      
\usepackage{lipsum}		
\usepackage{graphicx}

\usepackage{tikz}
\usetikzlibrary{plotmarks}
\usetikzlibrary{intersections,shapes.arrows}
\usepackage{pgfplots}
\pgfplotsset{compat=1.16}
\usetikzlibrary{patterns, positioning, arrows,spy,shapes.geometric,calc}
\usetikzlibrary{positioning, fit, arrows.meta, shapes}
\usepackage{cite}
\usepackage{amsmath,amssymb}

\usepackage{amsfonts}
\usepackage{pgfplots}

\newcommand{\Lagr}{\mathcal{L}}

\newcommand{\bs}[1]{\boldsymbol{#1}}

\title{Physics perception in sloshing scenes with guaranteed thermodynamic consistency}

\author{ {Beatriz~Moya}\\
	Aragon Institute in Engineering Research\\
	University of Zaragoza\\
	Zaragoza, Spain \\
	\texttt{beam@unizar.es} \\
	\And
	{Alberto~Badias} \\
	Aragon Institute in Engineering Research\\
	University of Zaragoza\\
	Zaragoza, Spain \\
	\texttt{abadias@unizar.es} \\
	\And
	{David~Gonzalez} \\
	Aragon Institute in Engineering Research\\
	University of Zaragoza\\
	Zaragoza, Spain \\
	\texttt{gonzal@unizar.es} \\
	\And
	{Francisco~Chinesta} \\
	ESI Group chair. PIMM Lab.\\
	ENSAM Institute of Technology \\
	Paris, France \\
	\texttt{francisco.chinesta@ensam.eu} \\
	\And
	{Elias~Cueto} \\
	Aragon Institute in Engineering Research\\
	University of Zaragoza\\
	Zaragoza, Spain \\
	\texttt{ecueto@unizar.es} 
}

\begin{document}
\maketitle

\begin{abstract}
Physics perception very often faces the problem that only limited data or partial measurements on the scene are available. In this work, we propose a strategy to learn the full state of sloshing liquids from measurements of the free surface. Our approach is based on recurrent neural networks (RNN) that project the limited information available to a reduced-order manifold to not only reconstruct the unknown information but also be capable of performing fluid reasoning about future scenarios in real-time. To obtain physically consistent predictions, we train deep neural networks on the reduced-order manifold that, through the employ of inductive biases, ensure the fulfillment of the principles of thermodynamics. RNNs learn from history the required hidden information to correlate the limited information with the latent space where the simulation occurs. Finally, a decoder returns data to the high-dimensional manifold, to provide the user with insightful information in the form of augmented reality. This algorithm is connected to a computer vision system to test the performance of the proposed methodology with real information, resulting in a system capable of understanding and predicting future states of the observed fluid in real-time.

\end{abstract}


\section{Introduction}
Simulating the world consists in the emulation of real dynamics in a virtual environment. Models within these simulations enable robots to experience perception of the world around them, thus allowing them to reason about the consequences of their actions \cite{liu2020role}. In this sense, fluid manipulation is a difficult task in AI-enabled robotics, and trustable physics-based simulation of liquids is desired for success  \cite{bates2019modeling} \cite{battaglia2013simulation} \cite{schenck2018spnets}.  

The merge between data-driven learning and knowledge about physics is becoming popular in dynamical modeling, enabling the study of complex systems of highly non-linear nature \cite{li2021kohn}. This conjunction is known as ``physics-informed deep learning''. Very different techniques may fall under this broad classification. Despite the common interest in introducing well-known physical knowledge into these approaches, there is a great divergence of proposals to be considered. From solving PDEs \cite{raissi2017physics} \cite{yang2021b} to learning constitutive laws based on invariants and conserved quantities \cite{liu2020ai} \cite{course2021weak}, these works establish a strong framework to work towards generalizable deep learning, extending models' capacity to learn about new situations out of the range of training. 

While these techniques keep improving, they face an important difficulty. In general, experimental campaigns to obtain the sets of governing variables are not always possible. Some works propose to model fluid dynamics with data coming from images \cite{bai2017data}  \cite{rodriguez20202d} or sensors \cite{bieker2020deep}, but that information could not be enough to form a physically consistent description. In contrast, some proposals try to reconstruct the missing dynamical data. Here, we propose an image-based method combined with physics knowledge to reconstruct the dynamics of the fluid from the free surface detected in video frames.

Information comes from camera recordings. This means that 30 to 120 snapshots per second will be available, and we expect a similar feedback rate of the model to provide the user with a smooth sensation. Under such stringent rates, encoding the information available to a low-dimensional manifold is mandatory. In this condition, we can establish a bridge between real systems with their digital twins \cite{sancarlos2021rom} \cite{moya2020physically}. 

The present work aims to develop a thermodynamically admissible model for fluid dynamics understanding and reasoning. The simulation engine is coupled with a computer vision system to build online digital twins of fluids. The performance of the loop must achieve real-time speed to guarantee trustable decision-making. 

The information we require in the proposed physics-informed description ---not only position, but also the velocity and stress fields and internal energy--- is inaccessible to a commodity depth camera. We do not consider here more sophisticated systems such as particle image velocimetry, for instance. We hypothesize that the knowledge of the internal variables we need for a complete description of the fluid will come from the study of the evolution of the free surface. We first train an autoencoder from full-field computational data coming from simulations. Then, we evaluate the information of the free surface in sequences to distill from its history the dynamical information needed to find a correlation with the latent manifold. We use a recurrent neural network (RNNs) to reconstruct the dynamical state and project it to a reduced-order space where simulations are performed to achieve real-time performance. On top of that, an augmented reconstruction is provided afterward, outputting not only the state of the full fluid volume, but also velocity, internal energy, and stress fields that were not accessible in the first place.

To guarantee the thermodynamic admissibility of the simulations, we learn from data a particular formulation of the dynamics based on the so-called General Equation for the Non-Equilibrium Reversible-Irreversible Coupling (GENERIC) formalism \cite{grmela1997dynamics}. GENERIC constitutes a generalization of Hamiltonian dynamics to dissipative phenomena. Under the scope of GENERIC acting as an inductive bias \cite{battaglia2018relational}, or learning constraint, during the learning procedure, we ensure the accomplishment of the basic principles of energy conservation and entropy generation, thus providing a physically sound learning framework.

The paper is structured as follows. Section 2 describes the state of the art and recent works in the field. Section 3 describes the problem in detail. Section 4 exposes the method, from the projection of the dynamics to a lower dimensional manifold to the learning algorithm and its connection with partial measurements. The training and implementation details are explained in section 5. Section 6 showcases the results obtained with real-world measurements. The paper finalizes with a discussion of the results and an evaluation of the future developments that can derive from this work. 

\section{Works in the field}

\subsection{Self-supervised estimation of dynamical states}

Labeling is an indisputable bottleneck for data-driven prediction. Thus, we need a framework to substitute the need for labeling with a deeper understanding of the information available. This has clear importance in robotics and visual perception and understanding based on images \cite{schenck2018spnets} \cite{nair2017combining} \cite{nava2021uncertainty} \cite{yan2020self}.
Dynamical modeling, and specifically fluid dynamics, also needs to address this lack of information when performing simulation and validation. Many internal variables employed in the descriptions are not easily measurable. Whereas some works use information obtained from images \cite{bai2017data}  \cite{rodriguez20202d} or sensors \cite{bieker2020deep}, many opt for reconstructing the dynamical internal state for an accurate description \cite{rao2021physics}. Dynamics typically use strategically placed sensors that acquire data to recover the full set of quantities from sparse observations \cite{callaham2019robust}  \cite{sun2020physics}. Several works employ deep neural networks (DNN) for this purpose. Erichson et al. \cite{erichson2020shallow} propose the use of shallow neural networks for reconstructing fluid flows. Lye et al. \cite{lye2020deep} estimate the unknown input parameters in turbulent flows from observables. 

Our contribution consists in a hybrid proposal between image and physics-based reconstruction using computer vision analysis.

\subsection{Deep learning incorporating physics priors}

Despite the use of self-supervised learning to compensate for the lack of unlabeled data, supervised methods still need a large database for learning a model. Few data could jeopardize the accuracy of the approximations. That is the case of black-box schemes, which do not succeed in learning global expressions and efficient generalizations in low data regimes \cite{lecun2015deep}. Inductive biases \cite{battaglia2018relational} guide the network to learn specific correlations in data. On top of that, they favor convergence and reduce error bounds \cite{Maier:2019aa}. As a result, less data is required, and the results are more realistic and accurate. 

Physics-informed deep learning is a current trend in artificial intelligence. Many approaches leverage theoretical knowledge to improve algorithm training \cite{raissi2017physics} \cite{lusch2018deep} \cite{ayensa2021prediction}. Hamiltonian (thus, conservative) systems are regular test benchmarks for these techniques \cite{jin2020learning} \cite{hesthaven2020rank}. Nevertheless, systems of utmost importance, such as those that involve Newtonian and non-Newtonian fluid dynamics, require beyond-equilibrium schemes. Thermodynamic neural networks constitute a framework that enables the study of any physical system, including those of inherent dissipative nature \cite{masi2020material}. Yu et al. \cite{yu2020onsagernet} apply the generalized Onsager principle to infer relationships among the state variables that define the system to ensure the fulfillment of that principle. GENERIC \cite{grmela1997dynamics} describes the changes over time of a set of state variables---that must characterize the energy of the system--- to model its evolution of energy and entropy. The dynamics can be fully described at a coarse-grained scale by learning the manifold of its time progression \cite{ibanez2017data}. This learning theory has been successfully applied to model rather different and complex behaviors  \cite{gonzalez2019thermodynamically} \cite{gonzalez2021learning} \cite{ghnatios2019data} \cite{moya2019learning}. In recent works, it has been coupled with DNNs to build the so-called Structure-Preserving Neural Networks (SPNN) \cite{hernandez2021deep} \cite{hernandez2021structure}. 

If we focus on the field of fluid simulation, DNNs are an extended tool for its emulating \cite{kim2019deep}. \cite{tompson2017accelerating} \cite{miyanawala2017efficient} apply CNNs to characterize 2D and 3D fluid dynamics. In the case of \cite{bukka2021assessment}, the authors distill the dynamics of unsteady flows with RNNs. Following the same spirit of the latter work, \cite{wiewel2019latent} employs specifically LSTMs networks in reduced order manifolds. Graph neural networks are becoming popular in this field \cite{sanchez2020learning}. Regarding physically informed deep learning, there are also works related to the study of fluids. \cite{mao2020physics} apply PINNs to high-speed flows. The work of \cite{gao2021phygeonet} presents an approach for learning PDEs from Physics-informed CNNs. 

There exist plenty of approaches \cite{sanchez2020learning} \cite{li2020visual} \cite{wu2015galileo} \cite{schenck2018spnets} in the context of scene understanding and interaction with fluids. In \cite{schenck2018perceiving}, the authors propose closing the learning loop with observations that help to correct the errors of the simulation. This work implements Navier Stokes in conjunction with the theory of Smooth Particle Hydrodynamics\cite{monaghan1992smoothed}. In contrast, we propose purely data-driven learning in a reduced order space imposing physical priors to achieve accuracy and generalization, as well as real-time performance. We believe in the use of the GENERIC formalism to discover fast and accurately the patterns of the dynamics for perception and interaction tasks. 

\subsection{Computer vision in fluids estimation}
Although the algorithm is trained with computational data, the final goal is to connect it with real liquids to close the perception loop. Detection and tracking of fluids, as well as containers, may be difficult if they lack texture. The measurements obtained are usually invalid or noisy because the surfaces are not Lambertian \cite{Koppal2014}. We are interested in the detection of fluids, particularly the free surface. \cite{schenck2016detection} \cite{schenck2018perceiving} propose the use of CNNs to perform tracking of the fluids. In the work of \cite{do2016probabilistic}, the authors propose an algorithm for filling level detection with RGD-D cameras. We propose an approach similar to the one presented in the work \cite{eppel2016tracing}. We convert the color image into a binary image in black and white to detect the color gradient that appears on the free surface.

\section{Problem description}

In this work, we propose a strategy to train simulators for physical scene understanding. Similar strategies have been developed in recent times, see \cite{wu2015galileo} for instance. However, in contrast with these works, we aim to develop a technique in which quantitative---and not only qualitative---information is given. This is of utmost importance if our perception system is to be employed in an industrial framework in the form of a digital twin of an asset, for instance. Thus, it is of primary importance to ensure that the learned simulator can make predictions that adhere to known basic principles of physics, such as the laws of thermodynamics.

We have considered the problem of fluid sloshing as a proof of concept for this analysis. This is in part due to its practical interest (to construct robots able to manipulate fluids, for instance) but also for its generality: sloshing is a (nonlinear) physical phenomenon that presents several interesting and challenging characteristics. Among them, a dissipative behavior. While there is a plethora of works devoted to learning conservative phenomena, based upon Hamiltonian or Lagrangian descriptions (see, among others, \cite{course2021weak} \cite{bertalan2019learning} \cite{greydanus2019hamiltonian} \cite{jin2020symplectic} \cite{toth2019hamiltonian} \cite{zhong2019symplectic}), very little has been investigated for learning dissipative phenomena.

We consider a system whose time evolution is described in terms of a set of state variables that ensure the full description of its thermodynamical state. In general, any physical system can be described at different levels (micro, meso, macro), that incorporate different information. At the molecular dynamics scale, physics is described by the positions and momenta of every molecule. While at this scale Newtonian laws apply and everything is conservative, the number of degrees of freedom, and also the time scale at which the phenomenon evolves, makes this type of description useless. Coarser levels incorporate less information, but also involve fewer degrees of freedom \cite{Espanol}. At this level, the effect of unresolved variables (those of lower levels not considered) on the evolution of resolved ones (those considered in the coarse level) introduces dissipation, by the celebrated fluctuation-dissipation theorem \cite{kubo1966fluctuation}.

The present approach models Newtonian, and also non-Newtonian (shear-thinning), fluids at a hydrodynamics level in a laminar regime. At this level, the state variables required for a full description of a (possibly non-Newtonian) fluid are position $\bs q_j$, velocity $\bs v_j$, internal energy $E_j$, and stress fields $\bs \tau_j$, according to the GENERIC formalism \cite{espanol1999thermodynamically}. Thus, for a fluid discretized into $M$ particles, the set $\mathcal S$ of state variables is
\begin{multline}\label{intrinsic}
\mathcal S=\{ \bs z =( \bs q_j, \bs v_j, E_j, \bs \tau_{j} , j=1,2,\ldots, M) \in (\mathbb R^3 \times \mathbb R^3 \times\mathbb R \times\mathbb R^6)^M \}.
\end{multline}

With the help of an RGBD camera, whose detailed description can be found in Section \ref{sec:10}, only partial measurements of these variables are observable. We are not interested in using complex laboratory equipment, such as particle imaging velocimetry for instance, which would greatly limit the scope of our approach. Instead, we only have access to a set of points of the fluid's free surface at each video frame, as will be detailed below. This severely limits the amount of information at our disposal and obliges us to develop a system able to recover the dynamical state. In this sense, our methodology is linked to (at least, partially) self-supervised methodologies. 

As a proof of concept, we implement the learning algorithm to reason about the physics of different fluids contained in a glass subjected to arbitrary movements. This phenomenon has been first reproduced computationally to simulate the evolution of a discretized fluid under different initial conditions. It has been computationally modeled and simulated with Smooth Particle Hydrodynamics \cite{monaghan1992smoothed}.  In those simulations, the geometry is not a parameter of the problem. We perform four simulations over different types of fluids, with four different initial velocities to trigger the slosh, but we consider the same volume of liquid every time. Therefore, the initial particle discretization is always the same, and valid also for the real liquid, since we fill the cup with the same volume of liquid. As a control test, we selected a liquid (glycerine) whose particle variations are not extremely chaotic. Nevertheless, the particles do experience variations in position, especially those of the free surface, which is the main interest in this work, and this evolution is to be captured by the learning algorithm proposed. Then, we evaluate the state of the fluid at discretized time steps for each simulation. The required state variables are evaluated at each particle and stored for each time step. As a result, we have a collection of snapshots that describe the state of the fluid for training of the networks. Afterward, the system must be able to predict the temporal evolution of a liquid under different, previously unseen, conditions with limited information. Each particle has a tag or an index. This indexation is employed to define de information vector in a specific order. Also, we work on a local coordinate system: the positions of the particles are referred to the center of gravity of the cup, placed at the bottom of the volume. Given a continuous discretization for a non-too-chaotic sloshing, we did not consider it necessary to force the autoencoder to be permutation invariant in this specific example. In this work, we mainly focus on the study of the dynamics of different types of fluids to perform accurate and physically sound emulations in real-time, and the distillation of information from only measurements of the free surface.

\section{Methodology and architecture} \label{sec:2}

\begin{figure*}
\centering
\includegraphics[width=0.75\linewidth]{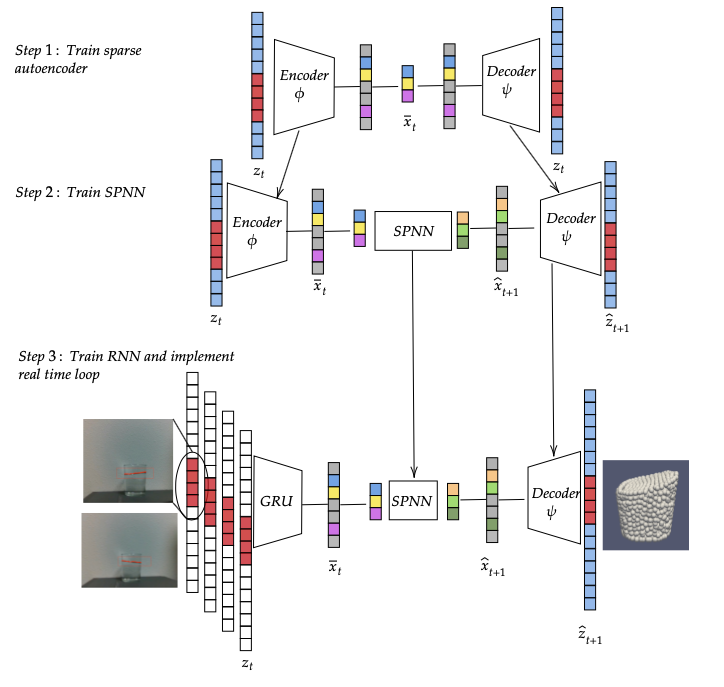}%
\label{fig_first_case}
\hfil
\caption{Sketch of the construction procedure for the deep neural network able to perceive sloshing phenomena. In a first step (first row), we must unveil the intrinsic dimensionality of the problem. To that end, we train a sparse autoencoder from computational full-field data. Once the number of variables governing the physics has been determined, we must train a structure-preserving neural network, able to integrate in time the state of the system. Thus, at step 2 in the figure, given the state of the system at time $t$, the decoder will output the state of the system at time $t+\Delta t$. But the main difficulty is that during runtime we do not have access to the high-dimensional state of the system, only a portion of it, represented in red in the input vector of the network. These values correspond to the position of the free surface of the fluid. Thus, the encoder must be substituted in step 3 by a GRU that takes the near history of the free surface to convert it into the reduced-order encoding of the system that will feed the thermodynamics-informed time integrator.}
\label{esquema}
\end{figure*}

The complexity of the algorithm just presented forces us to implement our system in three different steps, see Fig. \ref{esquema} for a graphical sketch of the implemented architecture. The highly dimensional nature of the problem motivates the reduction of the dynamics to carry out learning on an embedded space of a much lower dimension. In the case of learning and predicting new situations from real-world data, a correlation needs to be established between the data available (the free surface) and the latent space built from full computational descriptions. We hypothesize with the existence of features (distinctive attributes) in data sequences of the free surface that relates the history of partial measurements with internal variables of the fluid. An architecture based on recurrent neural networks can unveil those correlations, map the data acquired to the latent space, and output the reconstructed state in the next time step. 

To accomplish these requirements, we need to develop three different architectures. Firstly, we have to project our computational data to a lower dimensional manifold to train the algorithm efficiently and achieve real-time performance in the implementation. Secondly, we need to train a physics-informed integrator that will learn the evolution of the dynamics in the latent space. Finally, we work over an algorithm based on recurrent neural networks to substitute the encoder with a perceptron capable of projecting partial measurements to the latent space. It is worth mentioning that we chose to train in two different steps the autoencoder and the RNN and apply transfer learning. In transfer learning, we profit from the knowledge acquired in a new application. Sloshing dynamics is already a complex problem to apply model order reduction, and working with limited information from the free surface could complicate the process of learning a low dimensional manifold. Despite these considerations to optimize the algorithm, the need for deep neural networks that learn the patterns of the dynamics remains.

\subsection{Model reduction based on autoencoders}\label{sec:3}

In spite of the advances in terms of computational resources and calculation power, as well as the capacity of storage, model order reduction is still a necessary tool to deal with the abundance of data and the physical complexity of certain phenomena. Training a complex database could jeopardize the convergence to an optimal result. Such is the case of the database available to build the model of the present work. In addition, operating in a low-dimensional manifold could also foster the robustness and stability of the proposed learning scheme so that changes in the input do not strongly condition the stability of the solution. 

Currently, the development of these techniques is focused on capturing the important features of non-linearities. Fluid dynamics are led by strong non-linear structures hard to be learned by machine learning methods. New approaches arise from the perspective of autoencoders \cite{taira2020modal} \cite{erichson2019physics} as a preprocessing step to deal with turbulence and instabilities. 

Given a set of snapshots that live in a smooth finite-dimensional manifold $\mathcal M \in \mathbb R ^D$, with $D=13M$, as discussed in Eq. (\ref{intrinsic}), we aim to find a projection to a manifold of latent variables $\mathcal N \in \mathbb R ^d$ of much lower dimensionality, $d\ll D$. Autoencoders are an alternative to perform such reduction by finding patterns in the spatiotemporal structure of data. They are neural networks based on unsupervised learning which consist of two parts: an encoder that maps data to an embedded space, and a decoder that reconstructs latent information to the original space. The output of the decoder $ \hat{\bs z}_{t} = \hat{\bs z}(t)$ has to be equal to the input of the encoder $ \bs z_t = \bs z(t)$. These two structures are trained by backpropagating the error of the reconstruction,
\begin{align*}
\text{Encoder }\phi : \mathcal M \subset \mathbb R^D & \to \mathbb R^d\\
    									\bs z & \mapsto \bs x,
\end{align*}
\begin{align*}
\text{Decoder } \psi : \mathcal N \subset \mathbb R^d & \to \mathbb R^D\\
									\bs x & \mapsto \hat{ \bs z}.
\end{align*}

These schemes have shown good performance unveiling non-linear features to provide low and accurate representations. In this work, we decided to use sparse autoencoders (SAE) so the sparsity in the bottleneck distills the real low dimensionality of the problem. This is accomplished automatically by the sparse autoencoder by employing a L1-norm penalization \cite{ng2011sparse}. This can be interpreted, in the light of scientific machine learning, as a very practical form of imposing parsimony---in other words, Occam's razor---to the learned model \cite{liu2020ai}.

The backpropagated loss has two terms. With $N_{\text{snap}}$ as the number of snapshots introduced in the algorithm, the first loss term $\Lagr_{\text{mse}}^{\text{sae}}$ analyses the reconstruction error between the ground truth and the result of the decoder,
\begin{equation}\label{loss_MSE_SAE}
\Lagr_{\text{mse}}^{\text{sae}} =  \frac{1}{N_{\text{snap}}}\sum_{i=1}^{N_{\text{snap}}}(\bs z_i-  \hat{\bs z}_i)^2.
\end{equation}
 It is evaluated with the mean squared error (MSE) between the output and the input. Secondly, a regularizer term $\Lagr_{\text{reg}}^{\text{sae}}$ is introduced to enforce the sparsity of the solution for the latent state $\bs x_i$,
 \begin{equation}\label{loss_REG_SAE}
\Lagr_{\text{reg}}^{\text{sae}} =  \sum_{i=1}^{ N_d} | \bs x_i |,
\end{equation}
where $N_d$ is the dimension of the latent vector of the autoencoder. The size of the bottleneck is fixed a priori, and the number of non-vanishing entries (i.e., the intrinsic dimensionality of data) will be determined without user intervention during the training period.

The contribution of the regularization is weighted with a coefficient $\lambda_{\text{reg}}^{\text{sae}}$ to control its influence in the training process,
\begin{equation}\label{loss_SAE}
\Lagr^{\text{sae}} =  \Lagr_{\text{mse}}^{\text{sae}} + \lambda_{\text{reg}}^{\text{sae}}\Lagr_{\text{reg}}^{\text{sae}} .
\end{equation}

Although we have normalized the dataset, we decided to embed each group of state variables (position, velocity, internal energy, and stress tensor separated in normal $\bs \sigma$ and shear $\bs \tau$ components) separately to capture all the important features of each group of variables. In other words, we define five different autoencoders. The latent variables coming from each autoencoder are merged to form the latent space of the dynamics. This decision has been taken for optimization and accuracy reasons. The five encoders have the same structure. For all of them, the encoder and decoder have the same architecture, but inverted. They have been defined as fully connected and non-recursive layers that follow a feed-forward scheme.

\subsection{Learning the dynamical evolution based on Structure-Preserving Neural Networks}\label{sec:5}

We have already mentioned the fact that one of our primary interests is to develop a technique that satisfies known basic principles of physics. Thus, the learned time integrator should fulfill the first principles of physics to provide credible predictions of future events to help in decision-making. By structure-preserving neural networks (SPNN) we refer to a class of techniques that are constructed to satisfy some a priori known properties of the problem such as equivariance \cite{celledoni2020structure} or energy conservation \cite{greydanus2019hamiltonian} \cite{zhong2019symplectic}. In their most general form, they can be applied to conservative as well as dissipative problems, in which the principles of thermodynamics are satisfied by construction \cite{hernandez2021structure} \cite{hernandez2021deep}. In the case of SPNN, we work from a thermodynamical perspective to drive learning to admissible scenarios that will ensure the consistency of the results. For this purpose, we employ inductive biases that come from thermodynamic priors. 

When the phenomenon studied is dissipative, GENERIC is a particularly convenient formalism to describe its evolution in time. It presents a formulation to model the evolution of a vector of variables $\bs z$ from the analysis of the energy (Hamiltonian) potential in conjunction with a second (Massieu) potential that captures the dissipative nature of the dynamics. It is expressed in terms of the Poisson and friction brackets, which can be reformulated as matrix operators, as
\begin{equation}\label{GENERIC}
\frac{d\bs z}{dt}=  \bs L(\bs z)\frac{\partial E(\bs z)}{\partial \bs z}+\bs M(\bs z)\frac{\partial S(\bs z)}{\partial \bs z}.
\end{equation}

The product of the gradient of energy and the symplectic matrix, $\bs L \nabla  E$, models the conservative part of the time evolution, while the entropy gradient and the friction matrix $\bs M \nabla  S$ capture the occurring dissipative effects beyond equilibrium. To discern which part of the evolution is governed by conservative phenomena and which one by dissipation, an additional condition must be imposed, the so-called degeneracy conditions,
\begin{equation*}
\bs L\frac{\partial S}{\partial \bs z}=\bs 0, \;\;          \bs M\frac{\partial E}{\partial \bs z}=\bs 0,
\end{equation*}
that state that energy has nothing to do with dissipation (it is conserved in closed systems) and entropy is not responsible for reversible phenomena.

In fact, these degeneracy conditions guarantee the conservation of energy, 
\begin{equation} \label{energy_deg}
\dot{E}(\bs z)=\nabla E(\bs z) \cdot \dot{\bs z}= \nabla E(\bs z) \cdot \bs L(\bs z) \nabla E(\bs z) 
+\nabla E(\bs z) \cdot \bs M \nabla S(\bs z)=0, 
\end{equation}
and the production of entropy,
\begin{equation}\label{entropy_deg}
\dot{S}(\bs z)=\nabla S(\bs z) \cdot \dot{\bs z}= \nabla S(\bs z) \cdot \bs L(\bs z) \nabla E(\bs z) 
+\nabla S(\bs z) \cdot \bs M \nabla S(\bs z) \geq 0,
\end{equation}
respectively.

These requirements are guaranteed if we choose $ \bs L$ and $\bs M$ to be skew-symmetric and symmetric, positive semidefinite, respectively. For the sake of clarity, it is worth mentioning that GENERIC preserves these properties in the full-order as well as in the reduced-order manifolds, enabling the learning of its structure in the latent space. In other words, it works equally well for $\bs z$ and $\bs x$, provided that $\bs x$ expresses the dynamics with accuracy.

We work on a discretized context of data samples, and learning is performed over a discrete expression of GENERIC. In this case, we consider a Forward Euler approximation of the time derivative that describes the dynamical evolution of the system. From this expression, the discrete degeneracy conditions to be imposed can be straightly derived:
\begin{equation}
\frac{\bs x_{n+1}-\bs x_n}{\Delta t}=\mathsf L(\bs x_{n+1})\mathsf{ DE}(\bs x_{n+1}) + \mathsf M(\bs x_{n+1})\mathsf{ DS}(\bs x_{n+1}),
\end{equation}
where $\mathsf L$, $\mathsf M$, $\mathsf{DE}$ and $\mathsf{DS}$ represent the discretized versions of $\bs L$, $\bs M$, $\nabla E$ and $\nabla S$, respectively, and the subscript $n$ refers to time $t=n\Delta t$ and $n+1$ indicates time $t+\Delta t$, respectively.

The SPNN follows a feed-forward flow which consists of a set of fully-connected layers that learn the gradients of energy and entropy as well as the matrices $\mathsf L$ and  $\mathsf M$. The learning scheme enforces the skew-symmetry and symmetry of $\mathsf L$ and  $\mathsf M$ respectively and the degeneracy conditions. Given pairs of consecutive snapshots, the neural network learns the integrator of the dynamical problem. The input is the state vector of latent variables at time $t$ and $t+\Delta t$, $\bs x_{n}$ and $\bs x_{n+1}$. The output coming from the net is a vector that contains the predicted  $\mathsf L$,  $\mathsf M$, and gradients of energy and entropy of the current state. During runtime, the time integration is consecutively performed with these operators. 

The loss that carries the information to train and guide the SPNN is composed by two different terms. Firstly, the final output coming from the integration must match the ground truth. The accuracy is evaluated by measuring the mean squared error of these quantities,
\begin{equation}\label{loss_MSE_spnn}
\Lagr_{\text{mse}}^{\text{SPNN}} =  \frac{1}{N_{\text{snap}}}\sum_{i=1}^{N_{\text{snap}}}(x_i-\hat{x}_i)^2.
\end{equation}

Secondly, training is also governed by the degeneracy conditions. They are taken into account as a loss coming from the sum of the squared values of both conditions,
\begin{equation}\label{loss_de_spnn}
\Lagr_{\text{deg}}^{\text{SPNN}} =  \frac{1}{N_{\text{snap}}}\sum_{i=1}^{N_{\text{snap}}}(\mathsf L_i \mathsf {DS}_i)^2+(\mathsf M_i \mathsf {DE}_i)^2.
\end{equation}

Finally, the MSE loss is weighted with the hyperparameter $\lambda_{\text{mse}}^{\text{SPNN}}$ to control its influence in the global loss function of the network,
\begin{equation}\label{loss_RNN}
\Lagr^{\text{SPNN}} =  \lambda_{\text{mse}}^{\text{SPNN}} \Lagr_{\text{mse}}^{\text{SPNN}} +\Lagr_{\text{deg}}^{\text{SPNN}}.
\end{equation}

\subsection{Recurrent neural networks for state reconstruction}\label{sec:4}

The latent space is built from computational data obtained from simulations, for which we have a full description of the fluid states. Computational data includes internal variables important in our description, whereas they are unmeasurable by ordinary means from camera inputs. We hypothesize that, although internal variables are not measurable, their influence can be unveiled from the dynamical evolution of the free surface. From this information, we could analyze the features of the history of the dynamics where these internal variables will arise. 

Following the spirit of self-supervised learning, in the sense of unveiling not-given information, we propose an approach based on recurrent neural networks (RNNs) to reconstruct the state of the fluid. RNNs are structures that take into consideration the history of data working over sequences instead of learning from discrete and individual snapshots. RNNs are widely used, especially in the fields of natural language processing, speech recognition, or economics. Vanilla RNN often encounters, however, vanishing and exploding gradients \cite{pascanu2013difficulty}. Gated Recurrent Units (GRU)  \cite{cho2014learning} and Long Short-Term Memory (LSTM) units \cite{hochreiter1997long} are architectures capable of dealing with these problems. They include {\em gates}, or flows of information, to keep important information in long or complex sequences while forgetting irrelevant features. 

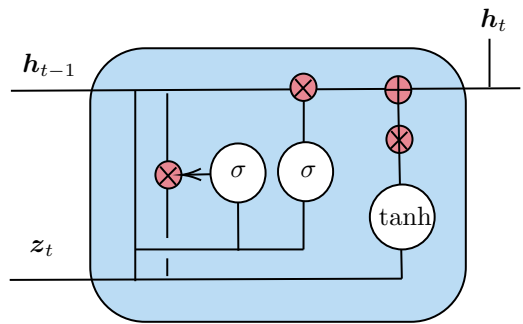
\begin{figure}[h!]
\centering

\tikzset{every picture/.style={line width=0.75pt}} 

\begin{tikzpicture}[x=0.75pt,y=0.75pt,yscale=-0.9,xscale=0.9]

\draw  [fill={rgb, 255:red, 80; green, 165; blue, 227 }  ,fill opacity=0.39 ] (91.01,92.77) .. controls (91.01,75.83) and (104.74,62.09) .. (121.69,62.09) -- (270.78,62.09) .. controls (287.72,62.09) and (301.46,75.83) .. (301.46,92.77) -- (301.46,184.82) .. controls (301.46,201.76) and (287.72,215.5) .. (270.78,215.5) -- (121.69,215.5) .. controls (104.74,215.5) and (91.01,201.76) .. (91.01,184.82) -- cycle ;
\draw    (46.53,85.94) -- (332,84.52) ;
\draw    (45.86,192.01) -- (265.61,191.3) ;
\draw    (116.23,85.94) -- (116.23,192.72) ;
\draw    (134.16,87.36) -- (134.16,168.52) ;
\draw  [fill={rgb, 255:red, 255; green, 255; blue, 255 }  ,fill opacity=1 ] (158.39,132.03) .. controls (158.39,122.89) and (165.3,115.48) .. (173.83,115.48) .. controls (182.35,115.48) and (189.26,122.89) .. (189.26,132.03) .. controls (189.26,141.17) and (182.35,148.58) .. (173.83,148.58) .. controls (165.3,148.58) and (158.39,141.17) .. (158.39,132.03) -- cycle ;
\draw    (116.23,174.92) -- (210.51,174.92) ;
\draw    (134.16,179.91) -- (134.16,189.87) ;
\draw    (210.67,85.41) -- (210.51,174.92) ;
\draw  [fill={rgb, 255:red, 232; green, 135; blue, 146 }  ,fill opacity=1 ] (127.52,133.1) .. controls (127.52,128.87) and (130.79,125.45) .. (134.82,125.45) .. controls (138.86,125.45) and (142.13,128.87) .. (142.13,133.1) .. controls (142.13,137.33) and (138.86,140.75) .. (134.82,140.75) .. controls (130.79,140.75) and (127.52,137.33) .. (127.52,133.1) -- cycle ;
\draw    (130.18,128.46) -- (139.03,138.67) ;
\draw    (129.95,138.2) -- (139.25,128.93) ;

\draw  [fill={rgb, 255:red, 232; green, 135; blue, 146 }  ,fill opacity=1 ] (203.2,83.98) .. controls (203.2,79.76) and (206.47,76.33) .. (210.51,76.33) .. controls (214.54,76.33) and (217.81,79.76) .. (217.81,83.98) .. controls (217.81,88.21) and (214.54,91.63) .. (210.51,91.63) .. controls (206.47,91.63) and (203.2,88.21) .. (203.2,83.98) -- cycle ;
\draw    (205.86,79.34) -- (214.71,89.55) ;
\draw    (205.64,89.08) -- (214.93,79.81) ;

\draw  [fill={rgb, 255:red, 255; green, 255; blue, 255 }  ,fill opacity=1 ] (196.23,131.32) .. controls (196.23,122.18) and (203.14,114.77) .. (211.67,114.77) .. controls (220.19,114.77) and (227.1,122.18) .. (227.1,131.32) .. controls (227.1,140.46) and (220.19,147.87) .. (211.67,147.87) .. controls (203.14,147.87) and (196.23,140.46) .. (196.23,131.32) -- cycle ;
\draw    (158.39,132.74) -- (144.13,133.06) ;
\draw [shift={(142.13,133.1)}, rotate = 358.75] [color={rgb, 255:red, 0; green, 0; blue, 0 }  ][line width=0.75]    (10.93,-3.29) .. controls (6.95,-1.4) and (3.31,-0.3) .. (0,0) .. controls (3.31,0.3) and (6.95,1.4) .. (10.93,3.29)   ;
\draw  [fill={rgb, 255:red, 232; green, 135; blue, 146 }  ,fill opacity=1 ] (256.32,85.41) .. controls (256.32,81.18) and (259.59,77.75) .. (263.62,77.75) .. controls (267.65,77.75) and (270.92,81.18) .. (270.92,85.41) .. controls (270.92,89.63) and (267.65,93.06) .. (263.62,93.06) .. controls (259.59,93.06) and (256.32,89.63) .. (256.32,85.41) -- cycle ;
\draw    (263.62,77.75) -- (263.62,93.06) ;
\draw    (256.32,85.41) -- (270.92,85.41) ;

\draw  [fill={rgb, 255:red, 232; green, 135; blue, 146 }  ,fill opacity=1 ] (256.98,113.17) .. controls (256.98,108.94) and (260.25,105.52) .. (264.28,105.52) .. controls (268.32,105.52) and (271.59,108.94) .. (271.59,113.17) .. controls (271.59,117.39) and (268.32,120.82) .. (264.28,120.82) .. controls (260.25,120.82) and (256.98,117.39) .. (256.98,113.17) -- cycle ;
\draw    (259.64,108.53) -- (268.49,118.73) ;
\draw    (259.41,118.27) -- (268.71,108.99) ;

\draw    (263.62,93.06) -- (265.61,191.3) ;
\draw  [fill={rgb, 255:red, 255; green, 255; blue, 255 }  ,fill opacity=1 ] (247.68,156.71) .. controls (247.68,146.65) and (255.81,138.5) .. (265.84,138.5) .. controls (275.87,138.5) and (284,146.65) .. (284,156.71) .. controls (284,166.77) and (275.87,174.92) .. (265.84,174.92) .. controls (255.81,174.92) and (247.68,166.77) .. (247.68,156.71) -- cycle ;
\draw    (173.83,148.58) -- (173.99,175.64) ;
\draw    (314.74,83.8) -- (314.74,56.75) ;

\draw (167.99,126.43) node [anchor=north west][inner sep=0.75pt]    {$\sigma $};
\draw (207.17,126.43) node [anchor=north west][inner sep=0.75pt]    {$\sigma $};
\draw (251.11,149.48) node [anchor=north west][inner sep=0.75pt]    {$\tanh$};
\draw (55.23,168.23) node [anchor=north west][inner sep=0.75pt]    {$\bs z_{t}$};
\draw (51.4,63.3) node [anchor=north west][inner sep=0.75pt]    {$\bs h_{t-1}$};
\draw (308.07,38.14) node [anchor=north west][inner sep=0.75pt]    {$\bs h_{t}$};

\end{tikzpicture}

\caption{Representation of a GRU cell. The three main paths indicated represent the update and reset gates, the new memory cell, and their connection to update the new hidden state transmitted to the next later.}
\label{gru}
\end{figure}

The basic idea behind the GRU architecture is to accumulate information from previous layers, see Fig. \ref{gru}. The hidden state $\bs h_t$ represents a summary of the features identified in previous sequences. $\bs g^{\text{update}}_t$ is the output of the update gate. This gate selects which information from the hidden state and the input sequence passes to the next step, modeled with a sigmoid activation function. In contrast, the reset gate $\bs r_t$ reflects the past information that should be avoided. A new memory cell $\bs n_t$, defined as the reset information, stores only the relevant information from the past. The output is the final hidden state $\bs h_t$ that accumulates the relevant information of past states and features learned from the current input sequence:
\begin{alignat*}{1}
\\
\bs g^{\text{update}}_t &= \sigma (\bs x_t U^z + \bs h_{t-1}W^z),\\
\bs r_t &= \sigma (\bs x_t U^r + \bs h_{t-1}W^r),\\
\bs n_t &= \tanh(\bs x_t U^h + (\bs r_t \bs h_{t-1})W^h),\\
\bs h_t &= (1-\bs g^{\text{update}}_t)\bs h_{t-1}+ \bs g^{\text{update}}_t \bs n_{t}.
 \end{alignat*}

GRUs have two main gates: an update gate, to update the hidden state, and a reset gate, which evaluates whether the previous cell state is relevant. In contrast with LSTMs, GRUs do not have a forget gate, which controls what is considered important to be remembered or to be rendered futile, and have fewer parameters. Despite having a simpler structure, GRU's performance is similar to LSTM in certain tasks. GRUs have proven to train faster and more efficiently with smaller datasets and shorter sequences \cite{chung2014empirical}.

The input of the network consists of a sequence of measurements of the position of the free surface of the fluid. Since it is firstly trained with computational data, we select the particles of the discretization that belong to the free surface at each time step. The batch of sequences is introduced in the network to pass through GRU recurrent layers doing a projection \textit{from-many-to-one}, i.e., introducing a sequence to obtain a single vector as output. The output vector of the GRU layers passes through a final forward fully connected layer with linear activation. The result of this process $\hat{\bs x}_t$ must match the latent state vector corresponding to the last snapshot $\bs x_t$ of the given sequence. The loss $ \Lagr_{\text{mse}}^{\text{GRU}} $ evaluates the MSE between the predicted latent state and the ground truth,
\begin{equation}\label{loss_MSE_RNN}
\Lagr_{\text{mse}} ^{\text{GRU}}=  \frac{1}{N_{\text{snap}}}\sum_{i=1}^{N_{\text{snap}}}(\bs x_i-\hat{\bs x}_i)^2.
\end{equation}

\section{Training and validation}\label{sec:6}

We train the algorithm from computational simulations performed in Abaqus CAE (Dassault Syst\`emes). It includes an add-on to perform conversion into particles and apply SPH in the simulations \cite{abaqus}. Each of the performed simulations is 2 seconds long, the time at which the fluid reaches the equilibrium state. We evaluate the state of the fluid at discrete time steps, equally spaced by time increments of 0.005 seconds. As a result, we have 1600 snapshots available for training. This dataset is split into two subsets: 80\% for training and 20\% for testing. The same training subsets are employed for training the three networks. The particles are labeled, and the data is stored for each time step into a vector following the order of the labeling. The geometry is not a parameter of the problem, and we have a constant discretization of the fluid. Nevertheless, the method is aimed to be able to learn the patterns of the dynamics, especially of those of the free surface, to emulate appropriately the sloshing perceived.

Once the three nets have been trained, we assemble the algorithm to feed the simulation loop with only a sequence of limited data. Instead of providing all the state variables obtained from the simulations, we pretend to only have the position at some points of the free surface. The sequence will be projected to the latent space, the SPNN will learn the dynamics, and the decoder will output results of the simulation augmenting the information originally given. The decoder provides the whole reconstruction of the fluid as well as the velocities, stresses and internal energy.

\subsection{Hyperparameters and characteristics of each net}\label{sec:7}

Each snapshot consists of a state vector of the position, velocity, internal energy, and stresses (shear and normal) evaluated at each particle of the discretized fluid. The fluid is discretized into 2134 particles. Thus, the global dimensionality is 27742. 

The global SAE is subdivided into five different SAE, one for each subset of state variables. The nets have been initialized following the Kaiming method in which, briefly speaking, the weight initialization follows a Gaussian, and biases are set to 0 \cite{he2015delving}. In addition, we apply linear activations in the first and last layers and ReLU for hidden layers. The optimizer chosen for these nets is Adam. A scheduler is programmed for updating the learning rate after 1000 and 3000 epochs. 

We adapt the architectures according to the input dimension and the complexity or noisy nature of the values, and encoder and decoder have a symmetric structure:

\begin{itemize}
\item Position: Input size is $D= 6402$ and output size $d= 20$. It is composed by $N_h = 2$ hidden layers of size $120$.
\item Velocity: Given the complexity of the velocity, we built a net of input size $D=6402$, output size $d=20$, $N_h=4$ hidden layers, and hidden size $200$. 
\item Internal energy: In the case of energy, input size is $D= 2134$, output size $d= 10$, and there are $N_h = 3$ hidden layers which consist of $40$ neurons each.
\item Normal stress: The normal stress tensor components are identical. Thus, the input shape of the net is $D= 2134$, the output shape is $d=20$, and it is composed of $N_h=3$ hidden layers of $200$ neurons.
\item Shear stress: This net had input size $D=6402$, $N_h=3$ hidden layers of $200$ neurons, and output size $d=20$.
\end{itemize}

The specific learning parameters such as learning rate ($lr$), weight decay ($wd$) and sparse weights ($\lambda^{\text{sae}}$) are defined for each SAE. Table \ref{table1} shows the parameters of each net. Given the complexity of the patterns learned, we require low learning rates.

\begin{table}[!t]
\renewcommand{\arraystretch}{1.3}
\caption{Training parameters for each SAE. \label{table1}}
\label{parameters2}
\centering
\begin{tabular}{c  c  c  c }
\hline
\bfseries   & \bfseries $lr$ &\bfseries $wd$ &\bfseries $\lambda^{\text{sae}}$ \\
\hline\hline
Position (q) &$10^{-4}$& $10^{-6}$& $10^{-3}$\\
Velocity(v) & $10^{-4}$& $10^{-5}$& $10^{-3}$\\
Internal energy (e) &$10^{-4}$& $10^{-5}$& $10^{-4}$\\
Normal stress ($\sigma$) & $10^{-4}$& $10^{-5}$& $5 \times 10^{-3}$\\
Shear stress ($\tau$)& $10^{-3}$& $10^{-6}$& $5 \times 10^{-3}$\\
\hline
\end{tabular}
\end{table}

After 10000 epochs, SAEs converge to optimal results. Taking into account the sparsity imposed to improve the reduction, the final dimension of each net is $d_{\text{position}}= 3$, $d_{\text{velocity}}= 3$, $d_{\text{energy}}= 2$, $d_{\sigma}= 3$ and $d_{\tau}= 2$. Thus, the final shape of the reduced space is $d_{\text{latent space}} = 13$. The latents obtained are the output of the RNN, and the input of the SPNN. This substantial reduction will not only reduce computing time and storage, but also improve the convergence to a solution in subsequent trainings. 

The recurrent neural network relates partial measurements of the fluid to the latent space that we have built. We consider that the only information accessible by ordinary means in real-time is information related to the free surface of the fluid. Specifically, we pretend we can only measure the position at some points of the free surface.

We take each full-order snapshot of the database to find the particles that represent the free surface of the liquid evaluating their height over their neighbors. Those points do not necessarily follow a balanced distribution since they are not uniformly distributed on the free surface. Given the density of the surface particle set, a simple linear interpolation is performed to obtain a uniform free surface grid. This step facilitates comparison between sequences in future predictions. As a result, we have equally discretized free surfaces of 21 points, which results in a vector of size 42. Despite reconstructing a 3D representation of fluids, we only consider a 2D data projection in the recurrent autoencoder to be consistent with the information that will come from the camera. Although we reconstruct the depth map of the images, depth measurements are still noisy. Thus, we decided to rely on 2D data, perpendicular to the depth, which represents the vertical movement of the liquid, that is more stable. Those coordinates represent the direction of the movement of the glass, and the vertical height of the liquid due to the sloshing effect.

Once we have the information related to the free surface at each time step, we assemble the sequences. We consider sequences of 16 snapshots. It is the minimum sequence size to guarantee that the features of the dynamics are correctly captured to find a mapping to the low dimensional manifold. With smaller sequences, the RNN does not learn good projections to the latent space. Since it is a short, although complex, sequence, GRU is optimal for this case. 

Even though the time step of the data in our database is 0.005 seconds, the camera streams depth measurements at a frequency of 60 Hz. Therefore, to assemble the sequences, we must choose snapshots equally spaced by approximately 0.015 seconds. 

The input size of the net is: batch size $\times$ sequence length $\times$ vector size. The net consists of three GRU hidden layers of 26 neurons, and there is a last feed-forward fully connected layer to connect the last GRU layer to the latent space of size 13. This last layer has linear activation. The optimizer for training this net is Adam, and parameters are set to $lr =10^{-3} $, $wd = 10^{-5} $. The learning rate is updated by a scheduler at 1000 and 3000 epochs.

We reach good results after 10000 epochs. Training loss is $1.19 \times 10^{-3}$, and test loss reaches a value of $2.3 \times 10^{-3}$.

The SPNN learns the integration scheme of the dynamics in the low manifold. The latent variables are the input of the net, so the input size is 13. Providing that $\mathsf L$ and $\mathsf M$ are skew symmetric and symmetric, respectively, we only learn the upper elements of the main diagonal. Thus, instead of learning the full matrices of dimension $d \times d$, we learn $d\cdot(d-1)/2$ elements for $\mathsf L$ and $d \cdot(d+1)/2$ elements for $\mathsf M$. Therefore, considering that the gradients have size $d$, the final output size is $d_{out}=d\cdot(d-1)/2 + d\cdot(d+1)/2 + d + d=195$.

We have reduced the complexity of the dynamics thanks to applying model order reduction. Despite it, the dynamics and the latent evolution are still complicated for training. We require a structure of $N_h=13$ hidden layers of size 195. 

The SPNN has been also initialized following the Kaiming method. We have applied ReLU activations for hidden layers, and linear activations in the first and last layers. The initial parameters are set to $lr =10^{-3} $, $wd = 10^{-5} $. The optimizer selected for training is also Adam, and the scheduler updates the learning rate after 1500, 2400, and 4000 epochs. Lower learning rates were required, again, due to the complexity of the features to be learned. The weight assigned to the MSE loss is $\lambda_{\text{mse}}^{spnn}= 10^3$, to give more importance to the reconstruction.

We train the SPNN for 5000 epochs. At that point, training and test losses are $3.2 \times 10{-3}$ and  $1.42 \times 10{-2}$, respectively. 

\subsection{Initial validation}\label{sec:8}

The performance of the method is tested with an input sequence extracted from the simulation of initial velocity 0.2 m/s. Given this sequence, the information is mapped to the latent state. Once we obtain the latent variables corresponding to that sequence, the SPNN integrates the dynamics until the fluid reaches the steady-state. We could provide information of sequences at each time step, but data acquisition is not always accessible in real scenarios (occlusion, connection problems, ...). In those cases, the simulation should continue until new information is provided. To this end, we test the ability of the method to continue integrating, and the stability of the simulation and results, by providing only one first snapshot. 

\begin{figure*}[!t]
\centering
\includegraphics[width=\linewidth]{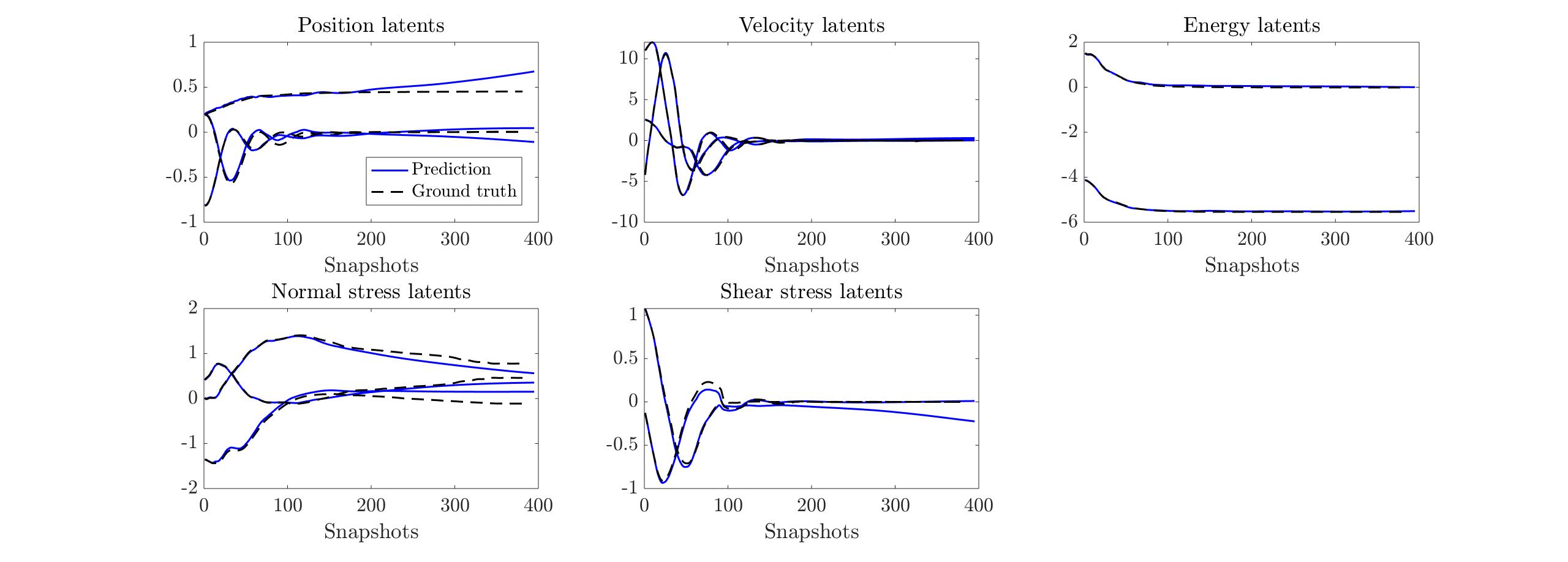}%
\label{fig_first_case2}
\hfil
\caption{Simulation results. Learning of the dynamics in the latent manifold. Dashed lines represent the time evolution of the latents that aimed to be emulated. Lines in blue represent the result of the SPNN in the latent manifold. }
\label{fig_sim}
\end{figure*}

\begin{figure}[h!]
\centering
\includegraphics[width=0.3\columnwidth]{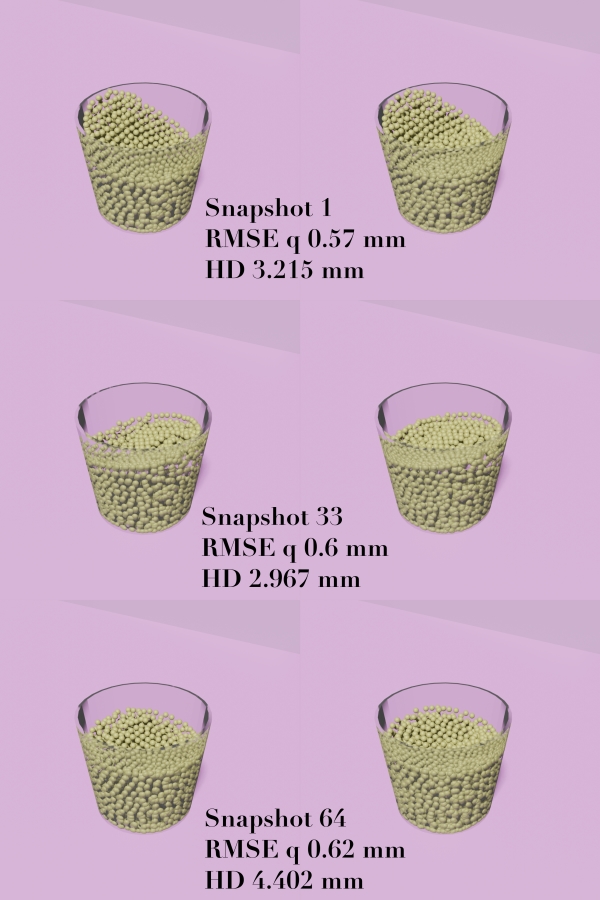}
\caption{Comparison of the reconstruction of the integration provided by the SPNN (right) with the ground truth (left). The selected snapshots correspond to peaks of the sloshing dynamics of glycerine. Specifically, we present the comparison for snapshots 1, 33, and 64 of the collection. The height of the cup is $7 cm$, and it is filled up to $5.6 cm$ approximately.}
\label{fig_sim4}
\end{figure}

\begin{figure}[h!]
\centering
\includegraphics[width=0.6\columnwidth]{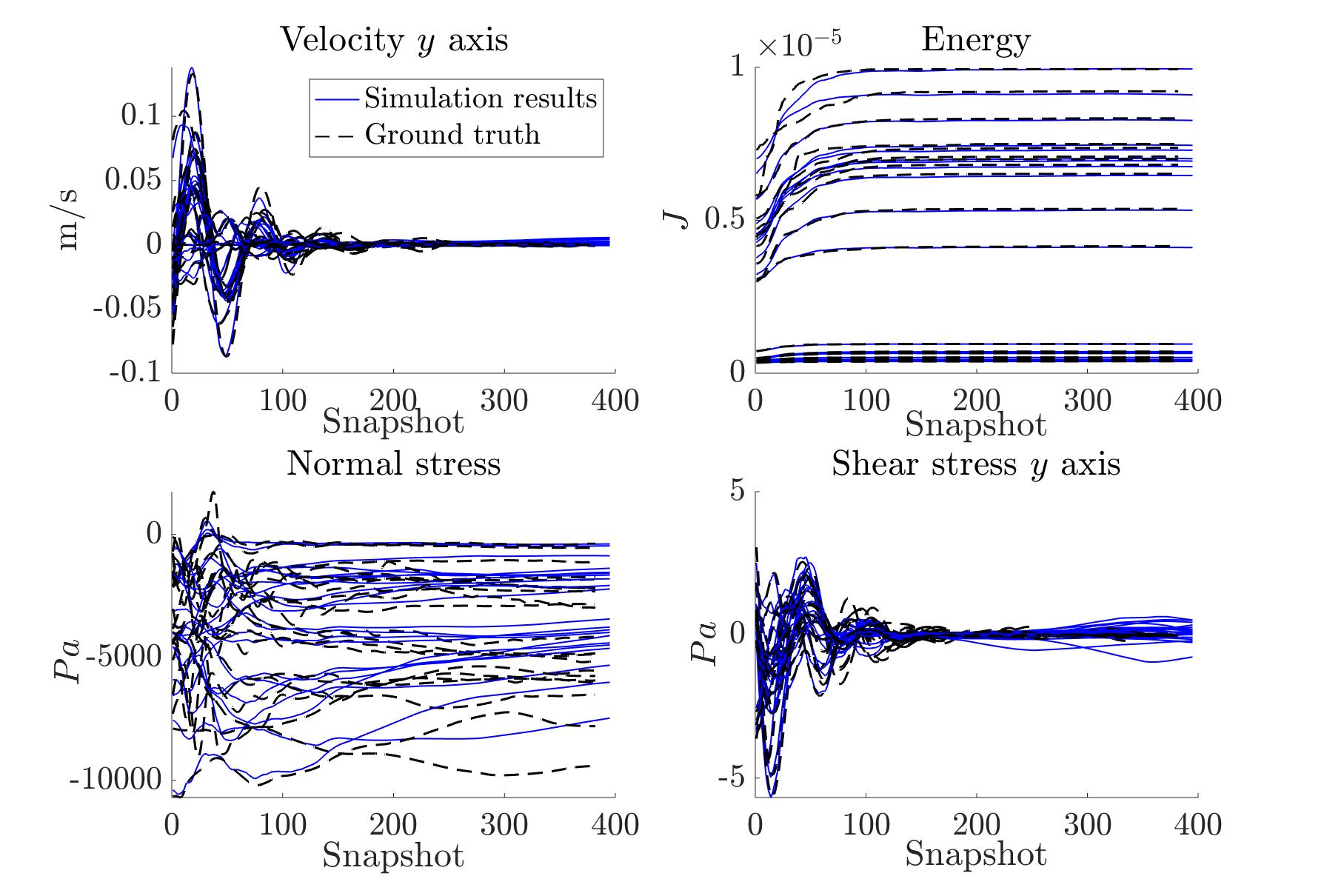}
\caption{Time evolution of selected state variables evaluated at 21 random particles. The graph shows a comparison between the simulated fields with the ground truth for the validation simulation of the algorithm.}
\label{computationalresults}
\end{figure}

Table \ref{comp_error} shows the MSE of the autoencoder proposed to reconstruct each group of state variables. These results have been compared with those obtained with POD\cite{ly2001modeling} taking 10 modes, and kPCA \cite{scholkopf1998nonlinear} with 4 modes. Modes are selected concerning the evolution of the eigenvalues obtained from each method. The AE achieves the same or improved levels of accuracy as POD and kPCA. Figure \ref{fig_sim} plots the simulation results in the reduced-order space. The initial state has been projected to the latent manifold to emulate the evolution of its behavior.  Fig. \ref{computationalresults} plots the time evolution of some state variables in the high-dimensional space for 21 randomly selected particles. Finally, Fig. \ref{fig_sim4} shows the comparison between the ground truth and the projection of the results to the high order manifold at three steps. This figure also includes the RMSE error of the reconstruction of each snapshot and the Hausdorff distance between the ground truth and the result. The Hausdorff distance (HD) \cite{huttenlocher1993comparing} evaluates the closeness of two sets by analyzing the largest distance between one set of points to another. If the HD is low, it resembles a high degree of similarity
$$
HD = \max\{\sup_{x \in X} d(x,Y),\sup_{y \in Y} d(y,X) \},
$$
being $X$ and $Y$ the two sets to be compared. In this context, the HD evaluates how well the shape of the result matches the shape of ground truth to have an indicator of the accuracy in the reconstruction. Thus, it compares the maximum ($sup$) distance from the ground truth to the output $\sup_{x \in X} d(x,Y)$, and vice versa $\sup_{y \in Y} d(y,X)$. Of these two distances, the maximum is the HD of the mismatch. After analyzing the results obtained in the computational phase, we decide to test the loop in a real scenario for the reconstruction of real fluids. 

\begin{table}[!t]
\renewcommand{\arraystretch}{1.3}
\caption{Loss comparison among SAE, kPCA and POD}
\label{parameters}
\centering
\begin{tabular}{c  c  c c}
\hline
\bfseries     & \bfseries Error AE&\bfseries Error POD &\bfseries Error kPCA \\
\hline\hline
$\bs q $& $0.149 \times 10^{-4}$ & $0.257 \times 10^{-4}$ &$0.141 \times 10^{-4}$ \\
$ \bs v$ & $4.1\times 10^{-4} $& $19 \times 10^{-4}$& $6.25 \times 10^{-4}$   \\
$e$ & $0.472 \times 10^{-4}$ & $0.64 \times 10^{-4}$ & $ 0.342 \times 10^{-4}$  \\
$\bs \sigma$ & $5.1 \times 10^{-4} $ & $ 20 \times 10^{-4}$ & $ 3.371 \times 10^{-4}$ \\
$\bs \tau$ & $ 0.798 \times 10^{-4} $ & $19 \times 10^{-4}$ & $3.36 \times 10^{-4}$ \\
\hline
\end{tabular}
 \label{comp_error}.
\end{table}

\section{Tests with real-world data}\label{sec:9}
Once the proposed strategy has been tested on computational data, it is extended to real-world problems. An RGBD camera is used to track the free surface of the fluid. This information is converted into sequences to predict the next state of the fluid. Finally, we not only have a reconstruction of the fluid, but also an augmented representation of it, by outputting the velocity, internal energy, and stress fields. 
\subsection{Data acquisition review}\label{sec:10}

\begin{figure}[h!]
\centering
\includegraphics[width=0.4\columnwidth]{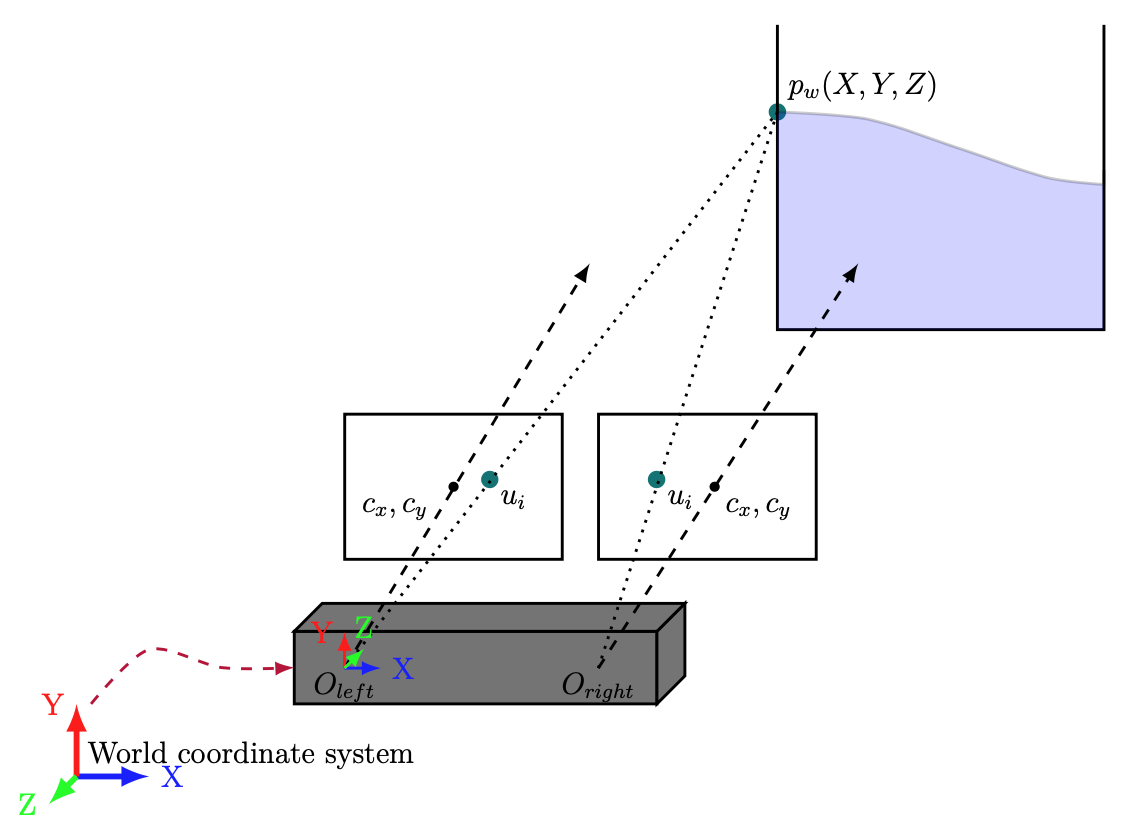}
\caption{Representation of the data acquisition system. The free surface is the tracked element of the algorithm to perform learning and the simulation of the dynamics.}
\label{fig_sim3}
\end{figure}

Firstly, we need to detect the free surface of the fluid and acquire the 3D coordinates to feed the algorithm with real data. We make use of a stereo camera for this purpose, although many of the presented results could equally be obtained with a standard camera. The model of our stereo camera is RealSense D415 (\url{https://www.intelrealsense.com/depth-camera-d415/}). This model of stereo camera provides both intrinsic and extrinsic parameters as well as depth measurements. Thus, the projection of the 2D image into the 3D world, and vice versa, is straightforward. The free surface is expressed in pixel coordinates $u,v$. The software provides the intrinsic parameters $f_x, f_y, c_x, c_y, s$, which are the focal length, the optical center coordinates, and the skew coefficient respectively. In addition to the intrinsic parameters $K$, we can also estimate (using Simultaneous Localization and Mapping techniques \cite{Mur-Artal:2017aa}), the rotation $\bs R$ and translation $\bs t$ components to complete the projection to the real world coordinates $X, Y, Z$ of the point $\bs p_w$,
\[s
\begin{bmatrix}
  u\\ 
  v\\
  1   
\end{bmatrix}
=
\begin{bmatrix}
  f_{x} & 0 & c_{x}\\ 
  0 & f_{y} & c_{y}\\
  0 & 0 & 1   
\end{bmatrix}
\begin{bmatrix}
  r_{11} & r_{12}  & r_{13}& t_{1} \\ 
  r_{21}  & r_{22} & r_{23}& t_{2}\\
  r_{31} & r_{32} & r_{33} & t_{3}
\end{bmatrix} 
\begin{bmatrix}
  X\\ 
  Y\\
  Z\\
  1   
\end{bmatrix},
\]
$$
\tilde{\bs x}_{s}=\bs K[\bs R|\bs t] \bs p_{w}.
$$
A sketch of the camera system is depicted in Fig. \ref{fig_sim3}. 

\begin{figure}[h!]
\centering
\includegraphics[width=0.27\columnwidth]{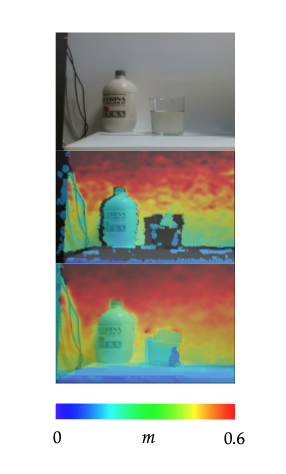}
\caption{Color and depth stream before (up) and after (down) applying the required filters to reconstruct the depth map of the streaming. }
\label{depth}
\end{figure}

It is worth mentioning that the original frame does not provide good enough depth measurements. It reports holes where the algorithm did not successfully measure the depth of the pixels. Measurements related to transparent objects are often invalid or noisy since their surfaces are not Lambertian, which is the main assumption of the measurement algorithm incorporated in the stereo camera. In other words, instead of reflecting light evenly in all directions, they also refract light, resulting in unmeasurable conditions for the technique defined. Our approach consists in applying some filters to enhance depth streaming. Firstly, we apply a decimation filter to reduce the complexity of the measurements to foster stability. Then, the frame is mapped to a disparity map where the spatial filter, to preserve the edges, and the temporal filter, to promote data persistency, are applied. This result is projected back to the depth map where the hole-filling filter is finally applied. The filtered depth map outputs a full depth field from which we can evaluate the position of the features of the glass and the free surface (see Fig. \ref{depth}). This procedure is fully detailed in the reference \cite{postprocessing}. 

\begin{figure}[h!]
\centering
\includegraphics[width=0.55\columnwidth]{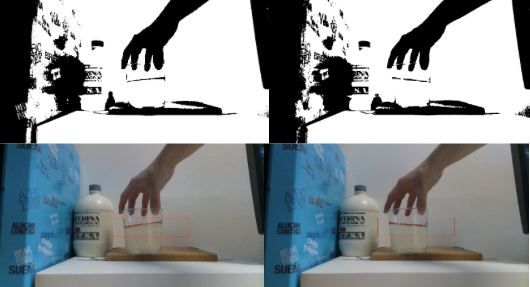}
\caption{Representation of the color frame and its conversion to a binarized image to seek the free surface. The area defined for searching is represented in the color frame as well as the points of the free surface detected in the black and white image.}
\label{bw}
\end{figure}

We binarize the color frame, also streamed by the camera, to convert the image into a black and white picture. The free surface appears as a gradient in the black and white image, see Fig. \ref{bw}. Since we prioritize the speed in the data streaming, we define an area for performing this analysis instead of forcing the recognition across the full image. The points of the free surface are detected, tracked, projected from frame coordinates to 3D, and stored. 

\subsection{Reconstruction and integration from video streaming}\label{sec:101}

We assemble the data obtained in the data acquisition step into sequences. These sequences feed the algorithm trained with computational data. The RNN projects the sequences to the latent space, the SPNN integrates the dynamics, and the decoder outputs the next state of the free surface and the position of the whole set of particles. Velocity, internal energy, $\sigma$, and $\tau$ evaluated at each particle are also provided. Therefore, we reconstruct the complete state of the fluid at the next time step only from the free surface. The video stream for validation consists of 800 frames, which is a recording of 12 seconds.

\begin{figure*}
\centering
\includegraphics[width=\linewidth]{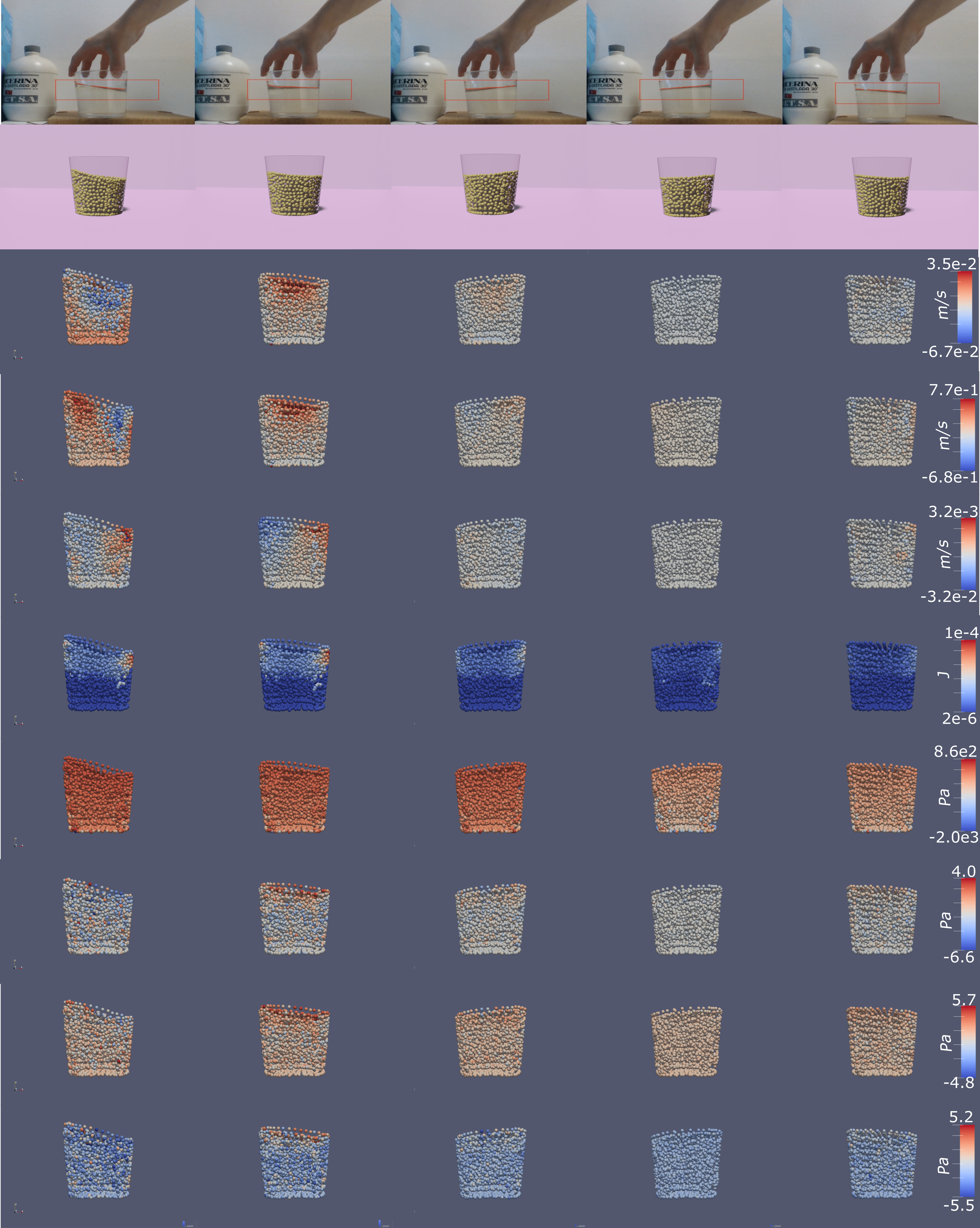}%
\label{fig_first_case3}
\hfil
\caption{Results for a 12 seconds video of a glass of glycerine. Eight snapshots of the sloshing sequence were selected for comparison. The selected snapshots have index 560, 565, 568, 572, 578 from left to right. The second row corresponds to the fluid reconstruction and prediction provided the previous snapshot. From row three to ten we show the additional information obtained from the reconstruction and simulation (velocity, energy and stress fields, respectively).}
\label{compare}
\end{figure*}

\begin{figure*}
\centering
\includegraphics[width=\linewidth]{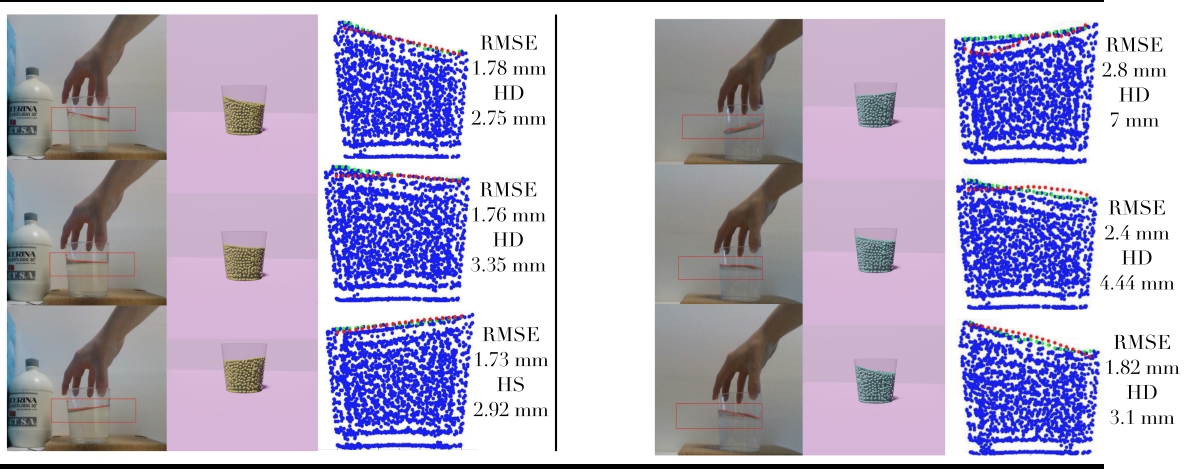}%
\label{fig_first_case4}
\hfil
\caption{Detail of the comparison of glycerine (left) and water (right) with the prediction. The third column of both liquids compares the predicted fluid volume (in blue), the free surface of the liquid volume (green) and the target free surface (in red). The RMSE and the Hausdorff distance (HD) that correspond to each snapshot are indicated.}
\label{compare}
\end{figure*}

Figure \ref{compare} shows the results of the algorithm compared to the real video streaming. We perform the reconstruction and integration over the whole sequence, i.e. no cuts were applied to the streaming and the method is applied continuously. We apply the three steps (RNN projection, integration, and decoding) over the full video in 3.42 seconds on an ordinary laptop (Macbook Pro 2013-3 GHz Intel Core i7), achieving (much more than) the real-time performance proposed. Some snapshots of the sloshing were selected and plotted in the first row of Fig. \ref{fig_first_case4}. The snapshots shown represent the peaks, which are the most critical states in manipulation, and some intermediate states between the peaks. The rest of the pictures correspond to the augmented information obtained with this method, which has been possible thanks to the physics-aware simulation framework. 

All results of the integration are stable, realistic, and close to the real result. We analyze objectively the results by evaluating the root mean squared error (RMSE) between the real $\bs y$ and the predicted $\hat{\bs y}$ free surfaces in $n$ snapshots of the video streaming. Ultimately, we feed the algorithm with the free surface in $t$ (from the video), and we compare the integration result ($t+1$) with the free surface in $t+1$ (from the video),

$$ 
\text{RMSE} = \sqrt{\frac{1}{n}\sum_{t=1}^{n}(\hat{\bs y}_i- \bs y_i)^2}.
$$

\begin{figure}[h!]
\centering
\includegraphics[width=0.4\columnwidth]{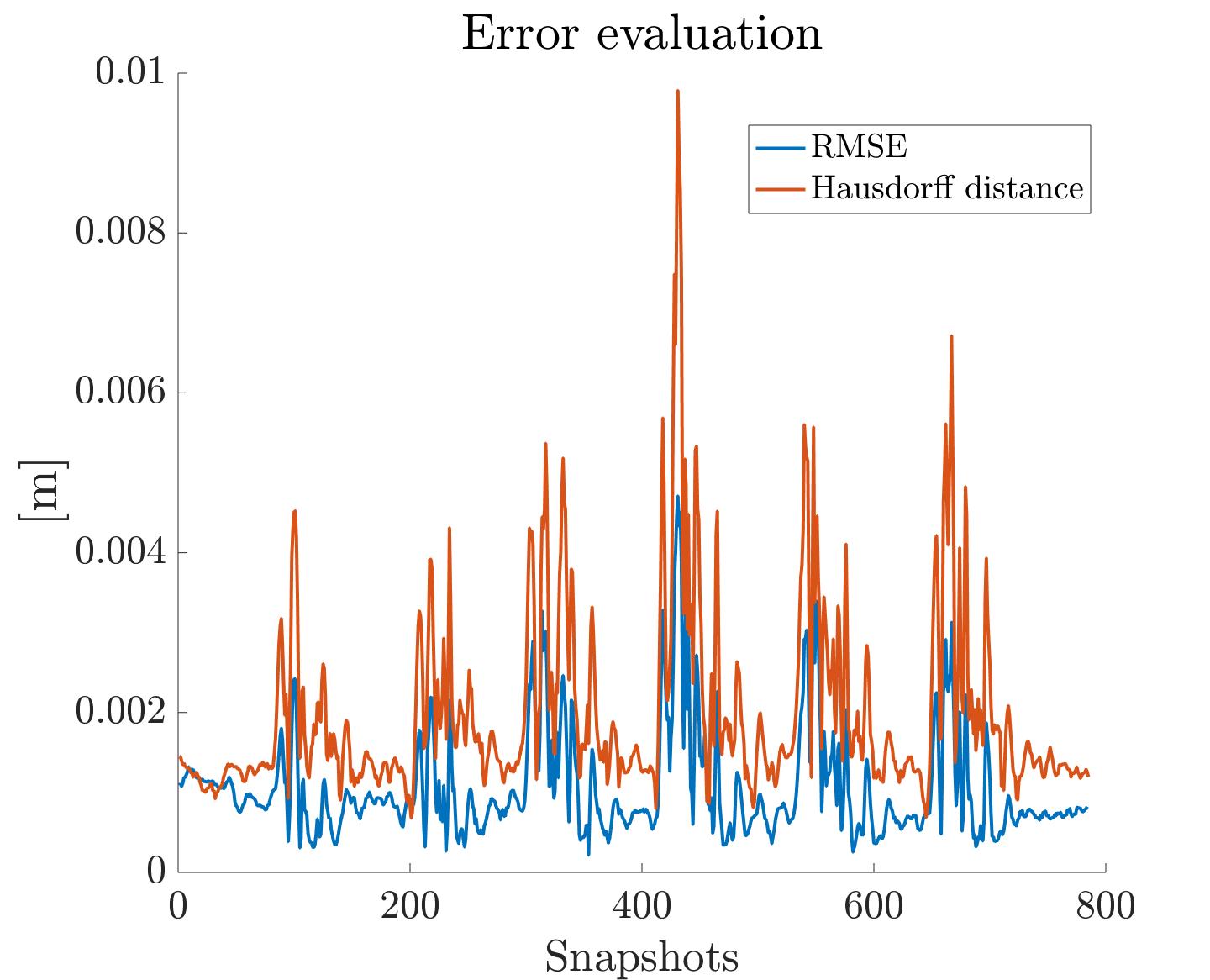}
\caption{Evolution of the mean squared error during the perception process of sloshing in a glass of glycerine.}
\label{mse}
\end{figure}

The evolution of the error along the video is represented in Fig. \ref{mse}. The error remains under 5 mm in the whole sequence of the length of the video and stays lower than 3 mm in the vast majority of it. We also evaluate the HD between the free surface that comes from the camera and the simulation. These results reflect the closeness between the free surfaces, for which there is not a larger deviation than 4--5 mm, even in the higher peaks of the sloshing. In some cases, higher deviations in the HD come from distortions in the detection of the free surface (like in the first snapshot of water presented).

\begin{figure}[h!]
\centering
\includegraphics[width=0.4\columnwidth]{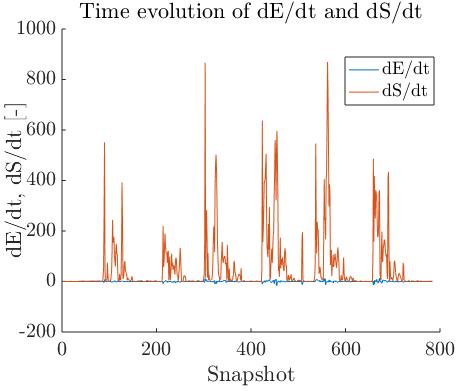}
\caption{Time derivatives of energy and Entropy along the video. The time derivative of Energy oscillates around zero, ensuring energy conservation. Entropy production is also ensured since the time derivative is always positive. }
\label{deds}
\end{figure}

Fig. \ref{deds} showcases finally the compliance of the principles of thermodynamics in the predictions. The time derivative of energy makes little oscillations due to the numerical approximation around 0, which means that we ensure the conservation of energy. In addition, the time derivative of entropy remains always positive, fulfilling its production.

\section{Conclusions and future work}\label{sec:11}

We have presented an approach for the physical perception of sloshing phenomena. It is based on physics-informed learned simulators connected to the real world employing commodity RGBD cameras. The algorithm has been trained with computational data to build a physically sound reduced-order manifold to learn the evolution of the dynamics dictated by the General Equation for the Non-Equilibrium Reversible-Irreversible Coupling (GENERIC). This thermodynamic framework ensures the physical consistency, accuracy, and realism of the results to promote informed decision-making. 

Provided the physics-aware approach for learning, only four simulations per fluid were needed to perform training that accurately mimics computational and real behaviors.  This approach has been completed with the development of a self-supervised technique to recover information of the dynamics that is unmeasurable by ordinary means. Advanced sensors and tools, such as PIV cameras, are available to evaluate data that cannot be extracted from a simple video stream. Nevertheless, we are still unable to measure important information for a full physical description. The suggested methodology fills in the gaps of information for the simulation of future dynamical states. 

Since the main purpose of the approach is to connect real-world systems with AI-guided simulators, we test the implementation of the integration scheme coupled with a data-acquisition system. The free surface is tracked by an RGBD camera, and the information obtained is used for fluid prediction. Notably, real-time is successfully achieved due to the reduction obtained through the application of autoencoders: 12 seconds of real-world time are analyzed in slightly more than 3 seconds, allowing performing decision making or using control algorithms. In addition, information is provided to the user by using augmented reality, i.e., by outputting the reconstructed fluid volume and a set of variables that may be important for decision-making on top of the video stream. 

Nonetheless, physics perception must achieve generality. It is unmanageable to train a model for each casuistic. Consequently, transfer learning must be extended to world and scene reasoning. Starting from a model such as the one proposed, incremental learning could set the base to extend learning and build hybrid twins that learn from evolving scenarios. Finally, the permutation-invariant condition is a desirable characteristic to work with unordered meshes \cite{qi2017pointnet++} \cite{wang2019dynamic} \cite{zhou2018voxelnet}. The consideration of this condition would help to achieve a higher degree of adaptivity and complexity previewed in future works. 

  \section*{Acknowledgments}

The work presented has been partially supported by the Spanish Ministry of Economy and Competitiveness through Grant number PID2020-113463RB-C31 and by the Regional Government of Aragon and the European Social Fund, research group T88. The authors also thank the support of ESI Group through the project UZ-2019-0060.

\bibliographystyle{unsrt}


\end{document}